\documentclass[10pt,twocolumn]{article}
\usepackage[a4paper,margin=0.8in,columnsep=0.25in]{geometry}
\usepackage[T1]{fontenc}
\usepackage{times}
\usepackage{microtype}
\usepackage{booktabs}
\usepackage{multirow}
\usepackage{float}
\usepackage{array}
\usepackage{amsmath}
\usepackage{amssymb}
\usepackage{tikz}
\usepackage{pgfplots}
\pgfplotsset{compat=1.17}
\usetikzlibrary{positioning,arrows.meta,fit,backgrounds,calc}
\usepackage[font=small,labelfont=bf]{caption}
\usepackage[hidelinks]{hyperref}
\hypersetup{pdftitle={SNO-PI: Label-Scarce SNOMED CT Entity Linking with Pre-trained Encoders and Curated Terminology Knowledge}, pdfauthor={Piyawoot Songsiritat}}
\usepackage[numbers,sort&compress]{natbib}

\newcommand{\concept}[1]{\textit{#1}}
\newcommand{\str}[1]{\texttt{#1}}

\title{\textbf{SNO-PI: Label-Scarce SNOMED~CT Entity Linking\\with Pre-trained Encoders and Curated Terminology Knowledge}}
\author{Piyawoot Songsiritat, MBBS, FRACGP\\The University of Queensland, Greater Brisbane Clinical School, Brisbane, Australia\\\texttt{p.songsiritat@uq.edu.au}}
\date{}

\begin{document}
\maketitle

\begin{abstract}
We present SNO-PI, the top-performing submission on SNOMED International's SNOMED~CT Entity Linking Benchmark as of
18 October 2026. SNO-PI achieves a macro character IoU of 0.5387, outperforming the
organisers' reference system (0.4582; based on the 2024 challenge winner's approach) by 18\% relative. Linking clinical mentions to
SNOMED~CT is a zero- and few-shot problem over a very large label space: the 272 annotated MIMIC-IV discharge
summaries cover only 6,595 of the roughly 219,000 eligible concepts, and a third of those occur only once. This
motivates a design built on transfer from pre-trained models and human-curated knowledge. SNO-PI combines
high-recall span proposal, SapBERT candidate retrieval and listwise ensemble reranking with three BERT-family
encoders, each further pre-trained with masked language modelling on unlabelled MIMIC-IV notes and then on
SNOMED~CT synonyms and hierarchy before task fine-tuning.
We find that curated knowledge helps most on concepts never seen in training, yet these concepts remain the main
source of error.
\end{abstract}

\section{Introduction}

Much of what clinicians record about a patient lives only in free text. A discharge summary carries diagnoses,
examination findings, procedures and allergies in the shorthand of clinical practice: ``hct'', ``SOBOE'',
``lap chole''. Until these are linked to a reference terminology such as SNOMED~CT, they are invisible to allergy alerts,
problem lists, registries and research queries. Unlike general-domain entity linking, clinical linking cannot rely on large labelled corpora: shareable clinical notes are
scarce because they contain protected health information, and labelling them requires clinical expertise, which
is costly and slow to obtain. The 2024 SNOMED~CT Entity Linking
Challenge~\citep{davidson2025} released a public corpus~\citep{hardman2024dataset} of discharge notes densely annotated with
SNOMED~CT~\citep{snomed}, drawn from MIMIC-IV-Note~\citep{johnson2023mimiciv,johnson2023note}.
Its organisers concluded that data, rather than model architecture, limited performance, with rare and unseen
concepts the hardest~\citep{davidson2025}. SNOMED International continues it as an open benchmark~\citep{benchmark2026}, hosted by
DrivenData and scored on a hidden set of about 20 notes.

The scale of the gap is stark. The 272 annotated notes contain about 75,000 labels, but they touch only 6,595 of the
roughly 219,000 concepts eligible for linking~\citep{hardman2024dataset}, or about 3\%. Any test note will therefore mention concepts for which no
labelled example exists. Yet the domain is unusually rich in \emph{curated} knowledge: SNOMED~CT itself supplies
synonyms, fully specified names and an is-a hierarchy for every concept, the UMLS~\citep{bodenreider2004umls} links
these to millions of further names, and hundreds of thousands of unlabelled clinical notes are available for
self-supervision.

SNO-PI is built around this asymmetry. Its contributions are:
(i)~a recall-first linking pipeline in which curated terminology enters at every stage;
(ii)~a staged training recipe (clinical MLM $\rightarrow$ SNOMED pre-training $\rightarrow$ task fine-tuning) that
lets cross-encoders learn concepts they never see labelled;
and (iii)~an analysis of where scarcity still bites, including note-level variation in annotation conventions.

\section{Task and the scarcity problem}

\paragraph{Task.} Systems output (\textit{note, start, end, concept}) tuples for mentions of concepts under three
SNOMED~CT hierarchies: \concept{Clinical finding}, \concept{Procedure} and \concept{Body structure}. Negated findings
and family history link to the plain concept, laterality is pre-coordinated into body structures, and lab
abbreviations map to measurement procedures (\str{hgb} $\rightarrow$ \concept{Measurement of total hemoglobin
concentration}). The primary metric is \textbf{macro} character IoU. Let $G_c$ and $P_c$ be the sets of
(note, character) positions covered by gold and predicted spans of concept $c$, and let
$\mathcal{C} = \mathcal{C}_{\text{gold}} \cup \mathcal{C}_{\text{pred}}$. Then
\begin{equation}
\mathrm{IoU}_c = \frac{|G_c \cap P_c|}{|G_c \cup P_c|}, \qquad
\text{Macro} = \frac{1}{|\mathcal{C}|} \sum_{c \in \mathcal{C}} \mathrm{IoU}_c ,
\label{eq:macro}
\end{equation}
so a rare concept counts as much as a common one and every spurious concept adds a zero. The secondary
support-weighted score weights $\mathrm{IoU}_c$ by the number of gold spans of $c$. Submissions are code packages that the organisers' harness executes
on hidden test notes, offline on one A100 within one hour, and scores with the official metric; participants never
see the hidden notes or their annotations.

\begin{figure}[t]
\centering
\begin{tikzpicture}
\begin{axis}[
  width=\columnwidth, height=4.6cm,
  xlabel={Annotated training notes}, ylabel={Share of test gold covered},
  xmin=0, xmax=280, ymin=0.2, ymax=1.0,
  xtick={25,50,87,130,174,204,272},
  xticklabel style={font=\scriptsize}, yticklabel style={font=\scriptsize},
  label style={font=\footnotesize},
  ytick={0.2,0.4,0.6,0.8,1.0},
  grid=major, major grid style={gray!20},
  axis line style={gray!60}, tick style={gray!60},
  legend style={font=\scriptsize, draw=none, fill=none, at={(0.98,0.04)}, anchor=south east},
  legend cell align=left]
\addplot[color={rgb,255:red,42;green,120;blue,214}, thick, mark=*, mark size=1.6pt]
  coordinates {(25,0.641) (50,0.730) (87,0.788) (130,0.825) (174,0.847) (204,0.859)};
\addlegendentry{gold spans whose text is a training alias}
\addplot[color={rgb,255:red,235;green,104;blue,52}, thick, mark=square*, mark size=1.5pt]
  coordinates {(25,0.423) (50,0.563) (87,0.656) (130,0.724) (174,0.771) (204,0.794)};
\addlegendentry{distinct gold concepts seen in training}
\addplot[color={rgb,255:red,42;green,120;blue,214}, thick, dashed, forget plot] coordinates {(204,0.859) (272,0.884)};
\addplot[color={rgb,255:red,235;green,104;blue,52}, thick, dashed, forget plot] coordinates {(204,0.794) (272,0.844)};
\end{axis}
\end{tikzpicture}
\caption{Label coverage grows slowly with annotation effort. Test gold coverage over random 20-note samples, for
random subsets of training notes (3 seeds). Dashed: log-linear extrapolation to all 272 notes.}
\label{fig:coverage}
\end{figure}
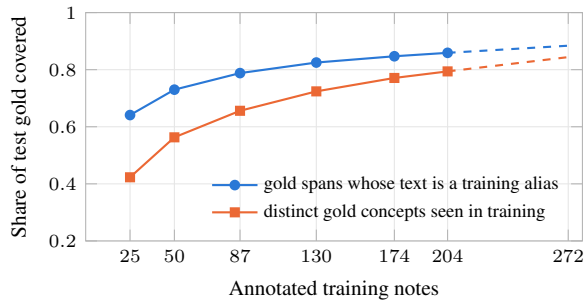

\paragraph{How scarce are the labels?} The 272 notes hold 75,463 spans over 6,595 distinct concepts. The
distribution is long-tailed: 2,504 concepts occur once and 4,788 five times or fewer. Figure~\ref{fig:coverage}
shows how slowly coverage grows with annotation effort. With 174 training notes, only 77\% of the concepts in a
random 20-note test sample have been seen at least once, and the curve extrapolates to only about 84\% at all 272
notes. Because the metric is macro-averaged, these unseen concepts (about 20\% of the concepts in a sample, but only
9\% of spans) weigh far more than their frequency.

\begin{figure}[t]
\centering
\begin{tikzpicture}[font=\footnotesize,
  line/.style={anchor=west, inner sep=1.5pt},
  ann/.style={draw=blue!70!black, fill=blue!10, rounded corners=1pt, inner sep=1.5pt, anchor=west},
  skip/.style={draw=gray!70, dashed, rounded corners=1pt, inner sep=1.5pt, anchor=west, text=black!75}]
\node[line] (h) at (0,0) {\textbf{ALLERGIES:}};
\node[ann]  (a1) at (0.3,-0.55) {Penicillins};
\node[skip] (a2) at (0.3,-1.05) {Sulfa (sulfonamide antibiotics)};
\node[skip] (a3) at (0.3,-1.55) {Codeine};
\node[draw=blue!70!black, fill=blue!4, align=left, text width=2.6cm, anchor=west, rounded corners=2pt]
  (c1) at (4.6,-0.55) {\concept{Allergy to penicillin}\\{\scriptsize SCTID 91936005}};
\draw[-{Latex[length=1.6mm]}, blue!70!black, thick] (a1.east) -- (c1.west);
\node[anchor=west, text=gray!90!black, align=left, font=\scriptsize] (lab) at (4.6,-1.3) {not annotated in this note\\(annotated in other notes)};
\draw[gray, dashed] (a2.east) -- (lab.west);
\draw[gray, dashed] (a3.east) -- (lab.west);
\end{tikzpicture}
\caption{Synthetic illustration of note-level variation: one list item is linked, its neighbours are not.}
\label{fig:allergy}
\end{figure}
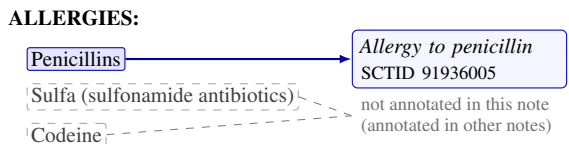

\paragraph{Scarce evidence of convention.} Six clinicians annotated the corpus, mostly one per note, with mean
pairwise Jaccard agreement of 0.85, which the organisers treat as an upper bound on achievable IoU~\citep{davidson2025}.
Within the remaining disagreement, peripheral choices vary between notes. In Figure~\ref{fig:allergy}, only one of
several listed allergies is annotated; clinically, this does not mean the patient tolerates the other drugs, but
reflects selective annotation, which matters when judging model errors against this reference. Such choices are
consistent within a note, plausibly reflecting individual annotators' habits: \str{pcp} is annotated at every
occurrence in 32 notes and at none in 38, and the same phrase elsewhere in the note predicts annotation better than
the phrase alone (83.5\% vs 70.9\%). Hand-written normalisation rules all lowered our score, whereas learned
rerankers absorbed these conventions implicitly through the \textsc{none} option and alias statistics.

\paragraph{Evaluation protocol.} We held out the 68 notes of the original 2024 test split plus 30 training notes
(dev) and trained on the remaining 174. Every score is the mean over 1,000 random 20-note subsets (sd about 0.014),
and a change was accepted only if it gained more than 0.003 on both dev and test on paired subsets. Final submissions were retrained on all 272 public notes. Models were
trained on NVIDIA GeForce RTX 50-series GPUs.

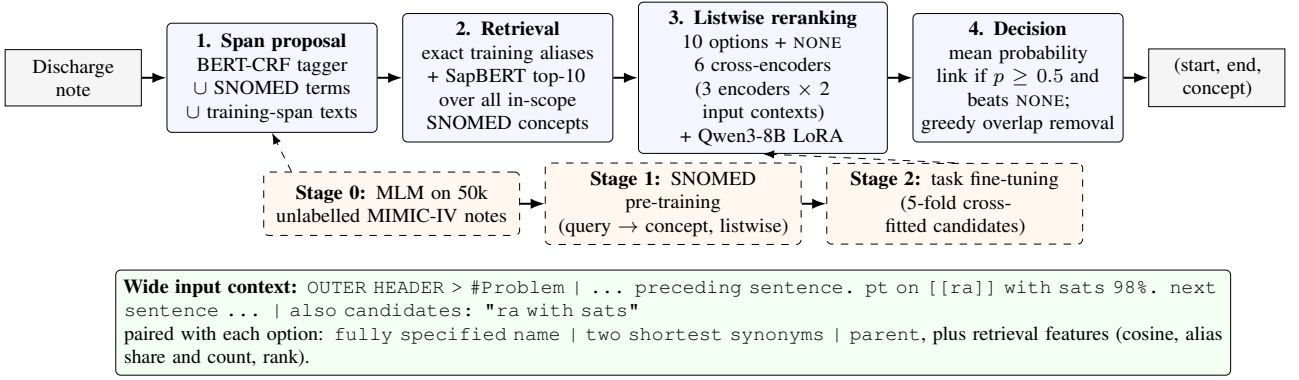
\begin{figure*}[t]
\centering
\resizebox{\textwidth}{!}{%
\begin{tikzpicture}[
  font=\footnotesize,
  stage/.style={draw, rounded corners=2pt, align=center, minimum height=1.55cm, text width=2.75cm, fill=blue!4},
  data/.style={draw, align=center, minimum height=0.8cm, text width=1.7cm, fill=gray!8},
  train/.style={draw, dashed, rounded corners=2pt, align=center, text width=3.4cm, minimum height=0.9cm, fill=orange!6},
  arr/.style={-{Latex[length=2mm]}, thick},
  darr/.style={-{Latex[length=1.6mm]}, dashed}]
\node[data] (note) {Discharge\\note};
\node[stage, right=0.35cm of note] (spans) {\textbf{1. Span proposal}\\BERT-CRF tagger\\$\cup$ SNOMED terms\\$\cup$ training-span texts};
\node[stage, right=0.35cm of spans] (ret) {\textbf{2. Retrieval}\\exact training aliases\\+ SapBERT top-10\\over all in-scope\\SNOMED concepts};
\node[stage, right=0.35cm of ret, text width=3.3cm] (rr) {\textbf{3. Listwise reranking}\\10 options + \textsc{none}\\6 cross-encoders\\(3 encoders $\times$ 2 input contexts)\\+ Qwen3-8B LoRA};
\node[stage, right=0.35cm of rr] (dec) {\textbf{4. Decision}\\mean probability\\link if $p \geq 0.5$ and\\beats \textsc{none};\\greedy overlap removal};
\node[data, right=0.35cm of dec] (out) {(start, end,\\concept)};
\draw[arr] (note) -- (spans); \draw[arr] (spans) -- (ret); \draw[arr] (ret) -- (rr);
\draw[arr] (rr) -- (dec); \draw[arr] (dec) -- (out);

\node[train, below=0.55cm of spans.south east, anchor=north, xshift=0.2cm] (mlm) {\textbf{Stage 0:} MLM on 50k\\unlabelled MIMIC-IV notes};
\node[train, right=0.35cm of mlm] (sno) {\textbf{Stage 1:} SNOMED pre-training\\(query $\rightarrow$ concept, listwise)};
\node[train, right=0.35cm of sno] (ft) {\textbf{Stage 2:} task fine-tuning\\(5-fold cross-fitted candidates)};
\draw[arr] (mlm) -- (sno); \draw[arr] (sno) -- (ft);
\draw[darr] (ft.north) -- (rr.south);
\draw[darr] (mlm.north west) ++(0.4,0) -- ($(spans.south)+(0.0,0)$);

\node[draw, rounded corners=2pt, fill=green!5, align=left, text width=15.6cm, below=0.35cm of sno, anchor=north] (input) {\textbf{Wide input context:}
\str{OUTER HEADER > \#Problem | \dots{} preceding sentence. pt on [[ra]] with sats 98\%. next sentence \dots{} | also candidates: "ra with sats"}\\
paired with each option: \str{fully specified name | two shortest synonyms | parent}, plus retrieval features (cosine, alias share and count, rank).};
\end{tikzpicture}}
\caption{SNO-PI architecture. Solid arrows: inference. Dashed boxes: the training recipe for the encoders. The
illustrative input is synthetic.}
\label{fig:arch}
\end{figure*}

\newpage 
\section{Compensating for scarce labels}

\begin{table}[t]
\centering\small
\setlength{\tabcolsep}{3pt}
\begin{tabular}{@{}>{\raggedright\arraybackslash}p{0.29\columnwidth}>{\raggedright\arraybackslash}p{0.32\columnwidth}>{\raggedright\arraybackslash}p{0.33\columnwidth}@{}}
\toprule
Knowledge source & Where it enters & Measured effect\\
\midrule
SNOMED terms and synonyms & span proposal, retrieval index, option text & only source of candidates for unseen concepts (R@10 .77)\\
\addlinespace
UMLS synonym sets (via SapBERT) & retrieval & R@10 .969 overall\\
\addlinespace
Pre-trained encoders (general and biomedical) & reranker initialisation, ensemble & prior that SNOMED pre-training builds on (SNOMED pre-training row); BioLinkBERT best on unseen in a controlled base-size comparison (.784)\\
\addlinespace
50k unlabelled notes (MLM) & tagger and encoders & tagger F1 +.009; small but consistent reranker gains\\
\addlinespace
SNOMED pre-training & reranker initialisation & concept acc.\ .26 $\rightarrow$ .85 / .71 (seen / unseen) with no labels (Table~\ref{tab:stages}); +.015 macro after fine-tuning (ablation)\\
\addlinespace
SNOMED hierarchy & scope filter, parent in option text, pre-training hard negatives & scope via is-a keeps the $\sim$22\% of gold annotations a semantic-tag filter would drop; part of SNOMED pre-training gain (+.015, above)\\
\bottomrule
\end{tabular}
\caption{Where non-label knowledge enters SNO-PI and what it bought (local test, 174 training notes).}
\label{tab:knowledge}
\end{table}

We describe the pipeline (Figure~\ref{fig:arch}) by the knowledge source each component draws on. Table~\ref{tab:knowledge} summarises the measured
effect of each source.

\paragraph{Curated terminology for recall.} Exact matches of in-scope SNOMED~CT terms are united with the spans of a
BERT-CRF~\citep{devlin2019bert,lafferty2001crf} tagger and with all span texts seen in training. The union reaches 94--95\% exact and 99\% overlap
recall of gold spans. Candidates are retrieved with SapBERT~\citep{liu2021sapbert}, a PubMedBERT~\citep{gu2021pubmedbert}
encoder trained on UMLS synonym sets, over an index of all in-scope SNOMED descriptions plus 12.7k training aliases.
Because the index is built from curated synonyms rather than from labels, it reaches concepts no annotator has used:
recall@10 is 0.97 overall and 0.77 on concepts unseen in training.

\paragraph{Pre-trained encoders as a knowledge base.} The rerankers start from encoders that already carry broad
language knowledge and, for two of the three, biomedical knowledge: Bio\_ClinicalBERT~\citep{alsentzer2019} was
trained on PubMed abstracts and MIMIC-III notes, BioLinkBERT-large~\citep{yasunaga2022linkbert} on PubMed with
citation links, and ModernBERT-large~\citep{warner2024modernbert} on a large general corpus including scientific
text. The staged recipe builds on this prior rather than replacing it. The clearest evidence is
Table~\ref{tab:stages}: after SNOMED pre-training alone, with no annotated notes, the encoder already picks the right
concept for 85\% of seen and 71\% of unseen gold spans. Biomedical pre-training matters most where labels are absent: in a controlled comparison of base-size encoders
under one early recipe, BioLinkBERT had the highest unseen-concept accuracy (.784, against .763 to .778 for the
others), and a general-domain encoder with a weaker adaptation (DeBERTa-v3~\citep{he2023debertav3}) was clearly
worse (.451 vs .487 macro). The three encoders were also chosen for diverse pre-training:
any two different encoders combined gain +.010 to .015 macro, while two Bio\_ClinicalBERT variants gain nothing.

\paragraph{Unlabelled clinical text.} The tagger and all three reranker encoders were further pre-trained with masked
language modelling~\citep{devlin2019bert} on 50,000 unlabelled MIMIC-IV discharge summaries (166M tokens;
held-out perplexity 13.94 $\rightarrow$ 2.86 for Bio\_ClinicalBERT~\citep{alsentzer2019}), a form of domain-adaptive
pre-training~\citep{gururangan2020} that adapts them to the
abbreviations, list structure and line-wrapped layout of discharge notes. In our experiments it improved both downstream tasks, span tagging and
reranking, consistently on dev and test.

\paragraph{SNOMED pre-training: learning concepts without labels.} Before seeing any annotated note, each reranker is
trained on SNOMED~CT itself: a synonym of a concept acts as the query, and the model must pick the right concept from
fully specified names, synonyms and parents among SapBERT-retrieved options, with the concept's parent and most
similar siblings added as hard negatives. This stage teaches the model the
terminology's structure across concepts that the 272 notes never mention, following the idea of enhancing
cross-encoders with knowledge-graph hierarchy for zero- and few-shot medical entity linking~\citep{gallego2025}. Table~\ref{tab:stages} shows the C-DAPT reranker after each
training stage. A separate ablation put the stage's gain after fine-tuning at +0.015 macro, almost all on unseen concepts.

\begin{table}[t]
\centering\small
\setlength{\tabcolsep}{4pt}
\begin{tabular}{@{}lccc@{}}
\toprule
 & \multicolumn{2}{c}{Concept acc.} & Linked\\
\cmidrule(lr){2-3}
Stage & seen & unseen & (unseen)\\
\midrule
SapBERT retriever (top-1) & .285 & .550 & .216\\
\midrule
C-DAPT, raw encoder & .262 & .266 & .000\\
\quad + SNOMED stage & .849 & .705 & .106\\
\quad + note fine-tuning & .948 & .829 & .725\\
\bottomrule
\end{tabular}
\caption{The C-DAPT reranker (Bio\_ClinicalBERT after MIMIC-IV MLM) after each training stage, local test. Concept
acc.: the gold concept is ranked first among the ten concept options, before the threshold and the \textsc{none}
check. Linked: it also passes the $p \geq 0.5$ threshold and the \textsc{none} check (unseen concepts only).}
\label{tab:stages}
\end{table}

\paragraph{Listwise reranking with curated option text.} Each span is scored against its 10 options plus an explicit
\textsc{none}, with a softmax over the 11 choices so that linking and abstention are learned jointly. Fine-tuning
minimises
\begin{equation}
\mathcal{L}=
\begin{cases}
-\log p_y-\lambda\log(p_y+p_{\varnothing}), & y\neq\varnothing,\\[5pt]
-\log p_{\varnothing}, & y=\varnothing,
\end{cases}
\label{eq:loss}
\end{equation}
where $p$ are the softmax probabilities, $y$ is the gold concept or $\varnothing$ (\textsc{none}) for non-gold
spans, and $\lambda > 0$ is a weight. The second term penalises
only probability on wrong concepts, so the model learns to abstain conservatively rather than link a wrong concept. Each option is
shown as its fully specified name, two shortest synonyms and its parent in the hierarchy, so the model can reason
about concepts by their curated description rather than by memorised labels. Each of the three encoders is fine-tuned twice with different input context,
giving six rerankers: a short context (the span's line under its nearest section header) and a wide context
($\pm$1 sentence under its section-header path, plus up to three \emph{competing overlapping spans},
\str{also candidates: \dots}). The competing spans in the wide context gained +0.018 dev / +0.013 test macro. The final submission adds a seventh member: Qwen3-8B~\citep{qwen3}, LoRA-tuned~\citep{hu2022lora} to
score all options in a single forward pass via a linear head on its hidden states, without text generation. Tested in
isolation, it is the strongest single member, and in the ensemble it gave a small improvement on the hidden test set
(0.5352 $\rightarrow$ 0.5387 macro).

\paragraph{Simulating scarcity in training.} Training candidates are built with 5-fold cross-fitted aliases and
dictionaries, so that about 20\% of training mentions look unseen, as they will at test time.

\paragraph{Decision.} Member probabilities are averaged; a span is linked if its top concept beats \textsc{none} with
$p \geq 0.5$, and overlaps are resolved greedily by confidence.

\section{Results}

SNO-PI ranks first on the benchmark on both metrics (Table~\ref{tab:leaderboard}).

\begin{table}[H]
\centering\small
\setlength{\tabcolsep}{3pt}
\begin{tabular}{@{}>{\raggedright\arraybackslash}p{0.68\columnwidth}cc@{}}
\toprule
System & Macro & Weighted\\
\midrule
\textbf{SNO-PI (rank 1)} & \textbf{0.5387} & \textbf{0.6629}\\
Rank 2 entry & 0.4985 & 0.6423\\
Rank 3 entry & 0.4695 & 0.6273\\
Rank 4 entry & 0.4657 & 0.6165\\
\textbf{Organisers' reference (Super Dictionary)} & \textbf{0.4582} & \textbf{0.6234}\\
Rank 6 entry & 0.4214 & 0.5831\\
Rank 7 entry & 0.4085 & 0.5621\\
\textbf{Organisers' reference (FAISS + Qwen3)} & \textbf{0.2321} & \textbf{0.2160}\\
\bottomrule
\end{tabular}
\caption{The public benchmark leaderboard on 18 October 2026.}
\label{tab:leaderboard}
\end{table}

Its macro score is 0.040 above the next-best entry and 0.081 (18\% relative) above the organisers' reference
Super Dictionary solution, which follows the approach of the 2024 winner, KIRIs~\citep{amit2024kiri}. Table~\ref{tab:history} traces the system's development; local estimates tracked the official scores.

\begin{table}[H]
\centering\small
\setlength{\tabcolsep}{3.5pt}
\begin{tabular}{@{}lcc@{}}
\toprule
 & \textbf{Local} & \textbf{Hidden test}\\
Development stage & dev / test & macro / wtd\\
\midrule
Three-encoder ensemble & .504 / .511 & .500 / .628\\
+ six members, wider context & .532 / .530 & --\\
+ retrained on all 272 notes & -- & .535 / .658\\
\textbf{+ Qwen3-8B member (final)} & -- & \textbf{.539 / .663}\\
\bottomrule
\end{tabular}
\caption{Development history; each row adds to the one above. The first row uses MLM and SNOMED pre-training. Local scores are the mean over 1,000 random 20-note subsets (trained on 174 notes).
Hidden-test scores are the official results of our submissions (October 2026).}
\label{tab:history}
\end{table}

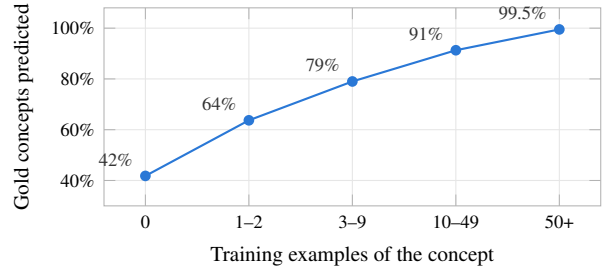
\begin{figure}[t]
\centering
\begin{tikzpicture}
\begin{axis}[
  width=\columnwidth, height=4.2cm,
  xlabel={Training examples of the concept}, ylabel={Gold concepts predicted},
  symbolic x coords={0,1--2,3--9,10--49,50+}, xtick=data,
  ymin=0.3, ymax=1.08, ytick={0.4,0.6,0.8,1.0},
  yticklabels={40\%,60\%,80\%,100\%},
  xticklabel style={font=\scriptsize}, yticklabel style={font=\scriptsize},
  label style={font=\footnotesize},
  grid=major, major grid style={gray!20},
  axis line style={gray!60}, tick style={gray!60},
  nodes near coords, nodes near coords style={font=\scriptsize, text=black!80, anchor=south east, xshift=-1pt},
  point meta=explicit symbolic]
\addplot[color={rgb,255:red,42;green,120;blue,214}, thick, mark=*, mark size=1.6pt]
  coordinates {(0,0.418)[42\%] (1--2,0.637)[64\%] (3--9,0.790)[79\%] (10--49,0.913)[91\%] (50+,0.995)[99.5\%]};
\end{axis}
\end{tikzpicture}
\caption{Concept prediction coverage increases with training frequency: share of test gold concepts predicted at least once, by the concept's
number of training examples (earlier three-encoder system, local test, 300 random 20-note samples).}
\label{fig:freq}
\end{figure}

\paragraph{Labels still matter most.} Curated knowledge narrows the gap but does not close it.
Table~\ref{tab:budget} replaces each stage with an oracle: a perfect reranker over our existing candidates would add
about 0.30 macro, against 0.06 for perfect retrieval and 0.12 for perfect spans. Concepts with 0--2 training examples
make up 36\% of concepts but 58\% of the remaining gap, and the share of gold concepts we predict at all rises
steeply with their first few labelled examples (Figure~\ref{fig:freq}).
 Unseen concepts mostly fail \emph{before} the reranker (no span or not retrieved),
while rare-but-seen concepts fail \emph{inside} it. Even a small increase in labels pays: aliases from 204 instead of
174 notes, with rerankers unchanged, added +0.004 test macro.

\begin{table}[t]
\centering\small
\begin{tabular}{@{}lcc@{}}
\toprule
 & Dev & Test\\
\midrule
Gold spans in span set (exact) & 95.2\% & 94.2\%\\
Retrieval recall@10 (seen / unseen) & .974 / .768 & .969 / .766\\
Linked to gold, seen concepts & .888 & .858\\
Linked to gold, unseen concepts & .678 & .685\\
Non-gold spans correctly rejected & 91.0\% & 91.6\%\\
\midrule
\textbf{Macro IoU, as is} & \textbf{.532} & \textbf{.530}\\
\quad + perfect reranker & .837 & .824\\
\quad + perfect retrieval & .891 & .882\\
\quad + perfect spans & 1.000 & 1.000\\
\bottomrule
\end{tabular}
\caption{Stage-wise analysis of the six-member system (174 training notes), with oracles applied cumulatively.}
\label{tab:budget}
\end{table}

\section{When more knowledge did not help}

Several attempts to inject further curated or pre-trained knowledge improved single models or intermediate metrics
but not the final ensemble (Table~\ref{tab:neg}). A recurring pattern is that gains outside SNOMED, or in recall
alone, do not transfer to the linking decision. The hierarchy near misses in Table~\ref{tab:neg} also show a limit of
exact-match scoring: predicting \concept{Renal calculus} for gold \concept{Staghorn calculus} scores zero, the same
as an unrelated diagnosis, although the two are clinically close. A hierarchy-aware metric would reflect clinical
usefulness better.

\begin{table}[t]
\centering\small
\setlength{\tabcolsep}{3pt}
\begin{tabular}{@{}>{\raggedright\arraybackslash}p{0.45\columnwidth}>{\raggedright\arraybackslash}p{0.51\columnwidth}@{}}
\toprule
Idea & Outcome\\
\midrule
\textbf{sap4}: our own SapBERT retrained on UMLS 2025AB with harder negatives, then mention-tuned on gold spans &
UMLS held-out acc@1 69.2\% $\rightarrow$ 80.8\%, but almost all of the gain is outside SNOMED, and harder negatives
\emph{lowered} unseen-concept recall. Mention tuning gave the best retriever measured (R@10 .955 $\rightarrow$
.962; unseen .738 $\rightarrow$ .785), yet only +.001 / +.002 in the ensemble.\\
\addlinespace
SNOMED is-a hierarchy encoder (HiT~\citep{he2024hit}) as retriever & Worse than SapBERT: R@10 .950 vs .955, unseen
.716 vs .738.\\
\addlinespace
Hierarchy-aware SNOMED pre-training loss (soft targets, relation margins); pre-training on every in-scope concept &
Within seed noise; unseen accuracy unchanged.\\
\addlinespace
\textbf{KRISSBERT}~\citep{zhang2022krissbert}: contextual retrieval as a second retriever, with source-agreement
features & Recovers 366 of 1,058 SapBERT misses (only 85 unseen); +.005 per three-model ensemble, $\leq$0 in the
six-model ensemble.\\
\addlinespace
Hierarchy near-miss resolution (parent/child pooling, specificity preference) & Oracle +.085; every rule $\leq$
+.0005, as annotators split a phrase between neighbouring concepts.\\
\bottomrule
\end{tabular}
\caption{A selection of negative results (paired $\Delta$ macro, dev / test).}
\label{tab:neg}
\end{table}

\section{Limitations}

The system used the AU 2025-08-31 release of SNOMED~CT, as the prescribed November 2025 International release could
not be obtained; 100 training concepts are inactive in it. With about 20 hidden notes, differences of a few
thousandths of macro IoU are within noise. Some error stems from note-level annotation
conventions unresolvable from the text.

\section{Conclusion}

With labels for approximately 3\% of eligible concepts, SNOMED~CT entity linking is a low-resource problem inside a
knowledge-rich domain. SNO-PI leads the benchmark by drawing on unlabelled clinical text, the curated synonyms and
structure of SNOMED~CT, and large pre-trained models at every stage, and this knowledge is most valuable precisely
where labels are absent. The steep gains from each additional labelled example, however, suggest that targeted
annotation of rare concepts remains the most effective next step.

\paragraph{Code availability.} The author plans to release the code on GitHub after clean-up at
\url{https://github.com/pisong314/sno-pi}.

\section*{Acknowledgements and disclosure}
The author holds the academic title of Senior Lecturer at the Greater Brisbane Clinical School, The University of
Queensland. This research was conceived, developed and conducted independently and was not funded or supervised by
the University; the views expressed are the author's own. This work used the MIMIC-IV-Note data under the PhysioNet Credentialed Health Data Use Agreement; no note text is
reproduced here beyond short, non-identifying span strings. The author designed the system and directed all
experiments. An AI assistant (Anthropic's Claude, via Claude Code) was used for coding and for drafting this
manuscript. The author reviewed all results and text and takes full responsibility for them.

\end{document}